\documentclass{llncs}
\usepackage{graphicx}
\usepackage{makeidx}  
\usepackage{tikz}
\usepackage{formatfig}
\usetikzlibrary{shapes,arrows}
\usepackage{todonotes}
\usepackage{amsmath,bm,times}
\usepackage{amssymb}
\usepackage[title]{appendix}
\usepackage{graphicx}
\usepackage{wrapfig, booktabs}
\graphicspath{{pics/}{figs/}}

\begin{document}
\title{A Cooperative Autoencoder for Population-Based Regularization of CNN Image Registration} 
\titlerunning{Cooperative Autoencoder Registration}
%
\author{Riddhish Bhalodia\inst{1,2} \and Shireen Y. Elhabian\inst{1,2} \and Ladislav Kavan\inst{2} \and Ross T. Whitaker\inst{1,2} }
%

\institute{Scientific Computing and Imaging Institute, University of Utah
  \and
 School of Computing, University of Utah
}
\maketitle              
\begin{abstract}
\vspace{-19pt}
Spatial transformations are enablers in a variety of medical image analysis applications that entail aligning images to a common coordinate systems. Population analysis of such transformations is expected to capture the underlying image and shape variations, and hence these transformations are required to produce {\em anatomically feasible} correspondences. This is usually enforced through some smoothness-based generic metric or regularization of the deformation field. 
%
Alternatively, population-based regularization has been shown to produce anatomically accurate correspondences in cases where anatomically unaware (i.e., data independent) 
regularization fail.  
Recently, deep networks have been used to generate spatial transformations in an unsupervised manner, and, once trained, these networks are computationally faster and as accurate as conventional, optimization-based registration methods. 
However, the deformation fields produced by these networks require smoothness penalties, just as the conventional registration methods, and 
ignores population-level statistics of the transformations. 
Here, we propose a novel neural network architecture that simultaneously learns and uses the population-level statistics of the spatial transformations to regularize the neural networks for unsupervised image registration. 
This regularization is in the form of a bottleneck autoencoder, which learns and adapts to the population of transformations required to align input images by encoding the transformations to a low dimensional manifold. 
The proposed architecture produces deformation fields that describe the population-level features and associated correspondences in an anatomically relevant manner and are statistically compact relative to the state-of-the-art approaches while maintaining computational efficiency. We demonstrate the efficacy of the proposed architecture on synthetic data sets, as well as 2D and 3D medical data.
\end{abstract}

%
%
%

\vspace{-20pt}
\section{Introduction}
\label{sec:introduction}
\vspace{-10pt}
Spatial transformations between sets of images play an important role in medical image analysis and are usually used for bringing distinct subjects into {\em anatomical correspondence}. This has many uses, such as the alignment of a population into a common coordinate system to compare functional/structural properties of specific anatomy, alignment of a new subject to an atlas, and in the study of anatomical shapes, where the transformations among and between images describe the morphology.
%
In all of these applications, there is an assumption, either explicit or implicit, that the ideal transformation should bring the images into an anatomical correspondence such that key parts of the anatomy are collocated in the transformed image(s). 
Some methods identify specific anatomical features 
and find transformations that ensure their alignment \cite{joshi2012diffeomorphic}.   Others find transformations that align unidentified image intensities/features, but {\em regularize} the problem with a smoothness penalty on the class of transformations \cite{beg2005computing,joshi2000landmark}. This approach has the advantage of potential generality, but it ignores known anatomical variability and correspondence. Thus, the {\em metric}, regularizations, or representations used to find these transformations do not incorporate any knowledge of transformations or class of transformations that best align members of a given population.   

Existing body of literature suggests that anatomical correspondences can be better learned (even in the absence of semantic/functional knowledge) in the context of {\em populations} of images or shapes \cite{grenander1991hands,cates2007shape,vialard2014diffeo}.   There is evidence that correct correspondence produces a population of transformations that is relatively easy to encode. 
This paper complements and extends these works by integrating population statistics (using non-linear models) into a deep neural network architecture for image registration, which we show is important for accurate characterization of anatomical correspondence.

Very recently, convolutional neural networks (CNNs) are utilized to regress coordinate transformations over the space of input images \cite{Vos2017EndtoEndUD,balakrishnan2018unsupervised}, in an {\em unsupervised} manner, by penalizing a metric of 
alignment between the input image pairs. These works are justified on the basis of computational speed or efficiency, as the feed-forward computation avoids non-linear, iterative optimization required for conventional image registration methods. However, CNNs for image registration offer other advantages, which are so far unexploited. In particular, CNNs do not rely on analytical representations of the coordinate transformation, the space of allowable transformations, or the optimization. This raises the possibility of incorporating empirical knowledge of the transformations, derived from a population of images, into the registration problem.  

In this paper, we propose using population-based learning of regularizations or metrics for controlling the class of transformations that CNN learns.    To achieve this, we introduce a novel neural network architecture that includes two subnetworks, namely \textit{primary} and \textit{secondary} networks, that work \textit{cooperatively}.   The primary network learns the transformations between pairs of images. The secondary network is a bottleneck autoencoder, that learns a low-dimensional description of the population of transformations, and {\em cooperates} with the primary network to enforce that the transformations adhere to a latent low-dimensional manifold. 

\vspace{-15pt}
\section{Related Work}
\vspace{-10pt}




Deformable image registration has been explored extensively, however,  challenges in generality, robustness, and efficiency remain. For brevity, we only focus below on the most closely related research.

%


Deformable registration is generally an ill-posed problem, and hence \textit{regularization} is required to achieve plausible transformations, avoid non-smooth transformations, and provide anatomically consistent results. Deformation fields are a classical way to represent transformations, typically regularized through smoothness penalty, usually in the form of Dirichlet/elastic penalty on the deformation \cite{bajcsy1989elastic}.  
For relatively low-dimensional representations, such as b-splines \cite{rueckert1999nonrigid}, the basis introduces a degree of smoothness, although some methods apply penalties on the b-spline coefficients. 
Diffeomorphic registration uses static or dynamic (with time-dependent velocity), smooth flow fields to represent the deformation while guaranteeing invertibility, and has been applied to image alignment and shape analysis \cite{beg2005computing}. The smoothness in the diffeomorphic setting is typically introduced as part of the metric on the flow field.
%


Recently, CNNs have been used for image registration to boost the computational efficiency by avoiding the non-linear, iterative optimization routines of conventional methods. 
Supervised methods for CNN training showed promising results \cite{krebs2017NRreg}, but this requires large amounts of labeled training data (i.e., registration examples solved with other techniques).
More recent work performs CNN-based registration in an unsupervised fashion  \cite{Vos2017EndtoEndUD,balakrishnan2018unsupervised}. The work of Balakrishnan et al. \cite{balakrishnan2018unsupervised} shows promising results on learning 3D brain registration displacement fields, improving the computational cost (after training) over the state-of-the-art traditional registration methods, such as ANTs \cite{avants2011ANTs}, while maintaining registration accuracy. Like most registration methods, this approach also uses smoothness on the deformation fields as a regularizer.

Early works by \cite{grenander1991hands} considered anatomical landmarks on a set of anatomical shapes, and suggested that anatomical variability is relatively low-dimensional. Later work used information-theoretic criteria to parameterize correspondences on populations of shapes \cite{cates2007shape}. Deformable transformations between images have also been confined to a low-dimensional representation that captures population characteristics \cite{rueckert2003automatic}. Statistical deformation models  \cite{rueckert2003automatic,joshi1997geometry} learn the probability distribution (subspace or manifold) of the deformation fields for a given population to reduce the dimensionality of the solution space and constrain the registration process. Low-rank representations and spatially varying metrics have also been proposed for diffeomorphic registration \cite{tanya2013lidiffeo,vialard2014diffeo}.    All these methods use linear models (e.g. PCA or low-rank correlations) to feed population statistics back into the registration process. In this paper, we introduce nonlinear models of the population and integrate these into a network architecture for registration.


This paper proposes a neural network architecture where one network influences another.  Few proposed systems of interacting neural networks include {\em generative adversarial networks} (GAN) \cite{goodfellow2014GAN} and its  variants, and domain adaptation (DA) \cite{ganin2016domain}. 
%
In these works, the primary network is {\em competing} with the secondary network as an adversary, and the steady states of these systems (in training) is a saddle point for the competing energies.  In the proposed work, the primary network is minimizing both its loss as well as the reconstruction loss of the secondary network, in an unsupervised setting---and thus we call these architectures {\em cooperative networks}.  
\vspace{-15pt}
\section{Methods}
\label{sec:methods}
\begin{figure}[!h]
    \centering
    \vspace{-8pt}
    \includegraphics[width=0.80\textwidth]{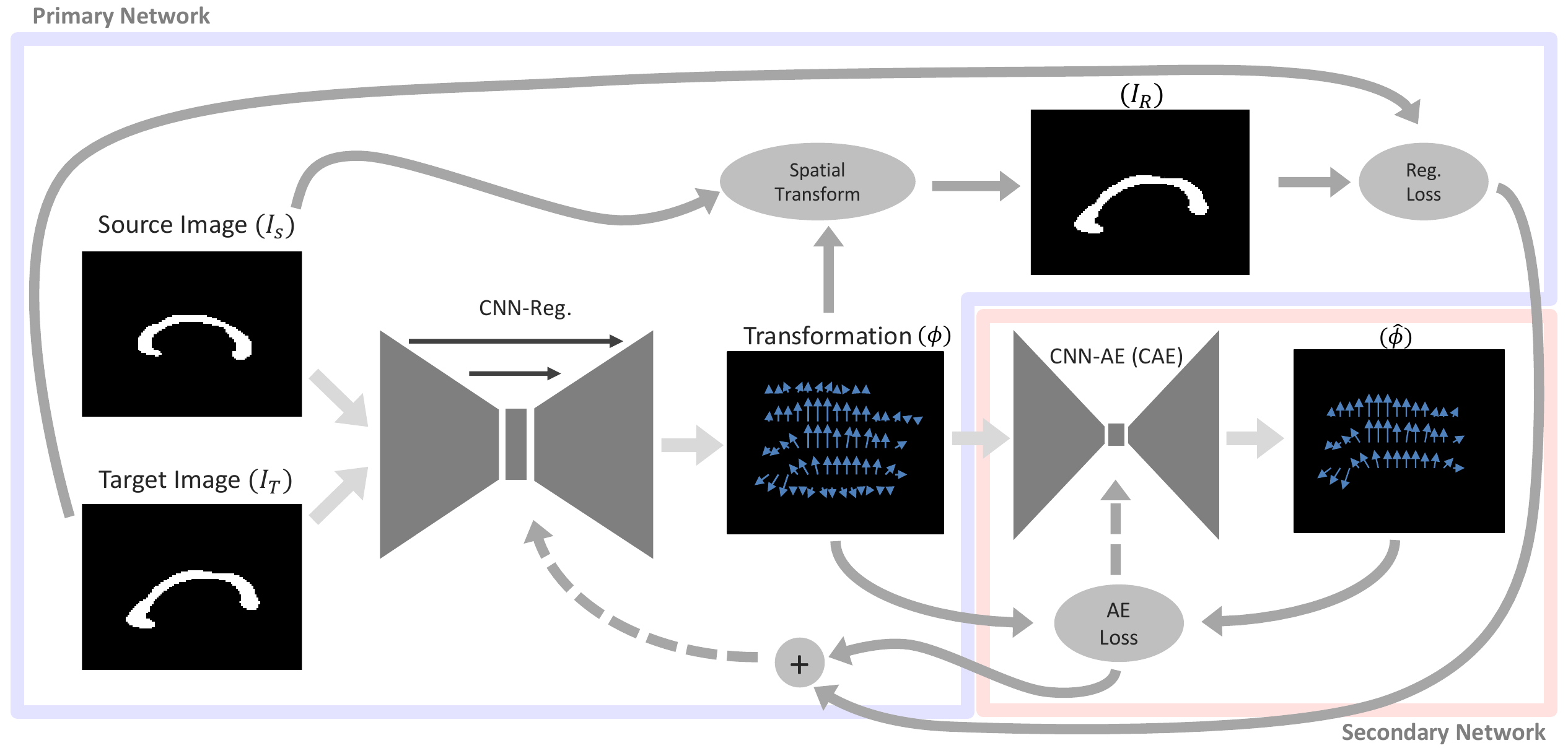}
    \vspace{-12pt}
    \caption{Cooperative network architecture, with the \emph{primary} unsupervised registration network depicted in the blue box, and the \emph{secondary} autoencoder based regularizer network in the red box.}
    \vspace{-17pt}
    \label{fig:arch}
\end{figure}

\vspace{-10pt}
The proposed cooperative network architecture is depicted in Figure \ref{fig:arch}. It consists of two interacting subnetworks, the \emph{primary} network aims at solving the primary registration task, and the \emph{secondary} network regularizes the solution space of the primary task. The architecture of the primary network is based on U-Net architecture (Figure \ref{fig:detarch}), in line with other registration approaches \cite{balakrishnan2018unsupervised}. Given a source ($I_S$) and a target ($I_T$) image pair (2D/3D), the network produces a displacement field $\phi$, corresponding to the warp that ideally should match $I_S$ to $I_T$. This displacement field, with the source image, is passed through a spatial transform unit \cite{jaderberg2015stunit} to produce a registered image ($I_R$). The primary network uses an image matching term between $I_R$ and $I_T$ as the loss function (e.g., $\mathbb{L}_2$ norm or normalized cross-correlation). To re-iterate, the displacement fields $\phi$ are not required for training, and hence, this is an unsupervised image registration architecture.

\vspace{-0.1in}
The secondary network is a bottleneck autoencoder, which we call a \emph{cooperative autoencoder} (CAE), that attempts to reconstruct the displacement field. The CAE's output is denoted as $\hat{\phi}$. The CAE is a CNN (Figure \ref{fig:detarch}) with an \emph{h}-degrees-of-freedom bottleneck layer (i.e. the latent space) represents the low dimensional nonlinear manifold on which the displacement fields should lie (approximately). We add the CAE's reconstruction loss ($\mathbb{L}_2$ loss given as $||\phi - \hat{\phi}||^2$) to the primary registration loss. CAE acts as a regularizer and pushes the network objective function so that it prefers, among many possible solutions, displacement fields that are accurately represented by the CAE.

The final objective function constitutes three terms (Eq. \ref{eqn:objective}). The first term represents the registration loss, the second term (weighted by $\alpha \geq 0$) is smoothness term \cite{balakrishnan2018unsupervised}, and, the third term (weighted by $\beta \geq 0$) is the CAE based regularization term. 
\begin{equation}
    \vspace{-3pt}
    \mathcal{Q} = Loss(I_T, I_R) + \alpha||\nabla \phi||^2 + \beta||\phi - \hat{\phi}||^2
    \label{eqn:objective}
\end{equation}

CAE training requires an initial set of transformation for a preliminary representation, hence, we start training with $\beta = 0$ (no CAE input), and a small smoothness with weight $\alpha$. We found that this length of initialization phase does not significantly affect the results of the system, and we always set it at 5\% of total iterations. After the initialization phase, we turn on the CAE and set $\beta$ to a non-zero value and $\alpha = 0$ (no smoothness), and train the primary and secondary network jointly (\emph{cooperatively}). 
\vspace{-15pt}
\section{Results}
\label{sec:Results}
\vspace{-10pt}
\setlength{\intextsep}{0pt}
\setlength{\columnsep}{5pt}
\begin{wrapfigure}[14]{r}{2.0in} 
  \begin{center}
    \includegraphics[scale = 0.40, trim = 0mm 0mm 0mm 25mm]{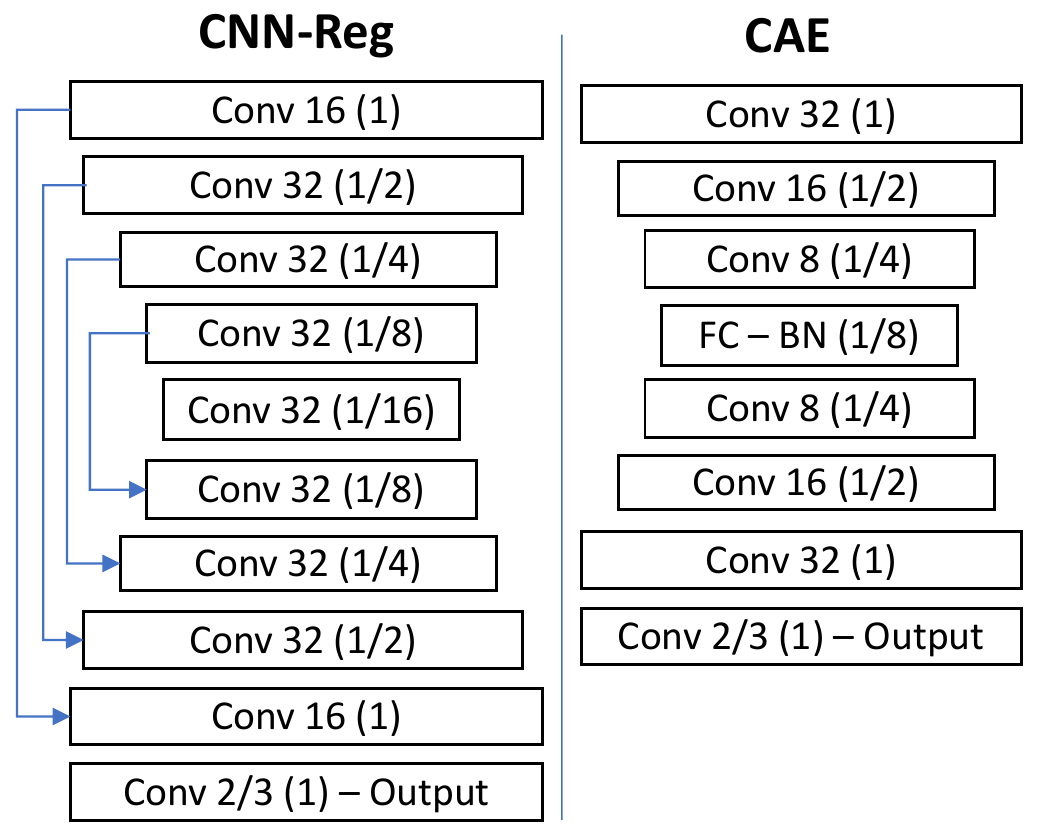}
    \vspace{-16pt}
    \caption{Left: primary network architecture (input: pair of images, output: displacement field between the images), which is then fed into the Spatial Transform (Figure \ref{fig:arch}). Right: architecture of the cooperative autoencoder.}
    \label{fig:detarch}
  \end{center}
\end{wrapfigure}
\setlength{\intextsep}{12pt plus 2.0pt minus 2.0pt}
\setlength{\columnsep}{10pt}
In this paper, we use the proposed method to register shapes, represented as binary images and/or distance transforms.  The same method applies directly to medical images. For each dataset, we train each network on \emph{all pairs} of images from the data, with random 25\% of the pairs set aside for testing. To clarify, this testing set is of completely held out pairs of images and the remaining 75\% of pairs is broken into training and validation set, Training on all pairs ensures that the CAE captures the inherent low-dimensional structure of the displacement fields while avoiding bias. However, the concept of cooperative networks is applicable to other training strategies (e.g. training with a given atlas image) or representations (e.g. momentum fields).

\noindent\textbf{Linear and Rotating Box-Bump}

\begin{wrapfigure}[19]{r}{3.3in} 
  \begin{center}
    \includegraphics[scale = 0.44, trim = 0mm 0mm 0mm 8mm]{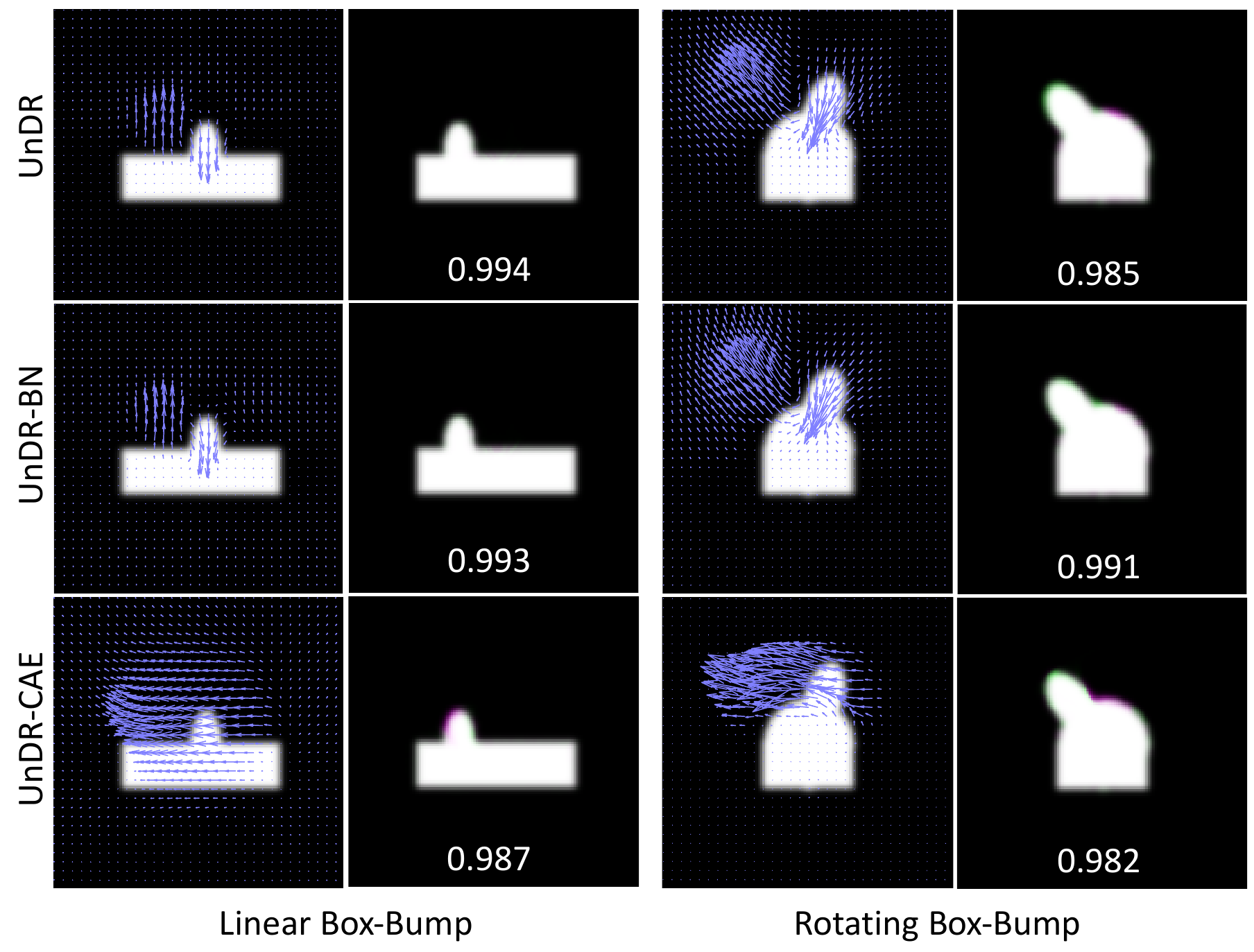}
    \vspace{-16pt}
    \caption{Linear \& rotating box-bump results with different methods, left figure shows the source with the field as produced by the network, and the right shows the false color difference image between the target and the registration output (white: correct overlap, green and magenta: mismatched pixels). }
    \label{fig:allBB}
  \end{center}
\end{wrapfigure}

Our first didactic dataset is a set of 2D {\em box-bump} (as in \cite{thodberg2003mdl}) images, where a protrusion on the surface of a rectangular shape is parameterized by its position along the side. We also use another synthetic dataset representative of rotational (non-linear) shape variations. Specifically, a protrusion is set atop of a circular base (parameterized by its angular position, between [-50, +50] degrees from the center). These linear and rotating box-bump datasets respectively represent a single linear and rotating (non-linear) mode of variation. We apply the proposed method on these datasets with the secondary network as cooperative autoencoder (CAE) with the bottleneck of dimension 1 and compare the resulting displacement fields with unsupervised deformable registration (UnDR) proposed in \cite{balakrishnan2018unsupervised}, which uses a smoothness penalty on the displacement fields and encodes no population-level information. We use $\mathbb{L}_2$ difference as primary loss, i.e. $Loss(I_R, I_T) = ||I_T - I_R||^2$. The results are shown in Figure~\ref{fig:allBB}, along with displacement fields and corresponding Dice coefficients, for a test pair of images.
We see that the registration accuracy measured using the Dice coefficient is comparable for UnDR and the proposed method (UnDR-CAE), but produces vastly different displacement fields. Cooperating networks capture a single transverse/rotating component for linear/rotating box bump, respectively, each derived from population statistics. In comparison, UnDR (for both datasets) compresses the protrusion for the source and expands it for the target, which correctly aligns the source and target shapes, but it does not discover the shape variation of the population. This is an important distinction: unlike UnDR, CAE leverages information about the population statistics of the data.

The core idea of cooperative networks is to restrict displacement fields to a low dimensional manifold. For comparison,  we also study some alternative strategies 
\setlength{\intextsep}{0pt}
\setlength{\columnsep}{5pt}
\setlength{\intextsep}{12pt plus 2.0pt minus 2.0pt}
\setlength{\columnsep}{10pt}
exploiting the same principle.  The first option is to reduce the latent space of the primary network architecture (UnDR) to a single dimension bottleneck, which we call ``UnDR-BN'', this represents a conventional alternative to the CAE.  The results for this approach are shown in Figure~\ref{fig:allBB} (UnDR-BN).
These results show that UnDR-BN is similar to UnDR, which can be explained, in part, by the {\em skip-connections} (Figure \ref{fig:detarch}) in the U-Net architecture used in UnDR.
An alternative to UnDR-BN architecture can be to introduce a $\mathbb{L}_1$ penalty on this layer to encourage sparsity. In our experiments, this leads to similar results as UnDR-BN, and for brevity, we do not present those results in this paper. We also provide additional results (in Appendix \ref{app:undr-bnwosc} ) with UnDR-BN, but with skip-connections of the U-Net architecture removed.

We hypothesize that cooperative networks can discover meaningful correspondences of shape, to validate we define landmarks (analytically) on the family of box-bump shapes (in
correspondence with the bump movement) and we evaluate how well each
method aligns these ground truth correspondences ({\em Landmark error}
in Table \ref{tab:proposed-results}), along with Dice coefficients
measuring registration accuracy. The computational cost of discovering displacement fields for a given image pair (testing step), are similar for both UnDR and the proposed method, i.e. CAE does not lose any of its speed over UnDR (speed is the main advantage of UnDR \cite{balakrishnan2018unsupervised}). UnDR-CAE registers with
similar accuracy as UnDR (measured by Dice coefficient), but
consistently achieves lower landmark errors due to the secondary
network which learns population statistics. It is also interesting to see the latent space variations as discovered by the single dimension of CAE and the additional results for this is provided Appendix \ref{app:latentspace}.

 For the CAE, we report the reconstruction error ($\frac{||\phi - \hat{\phi}||_{\mathbb{L}_2}}{||\phi||_{\mathbb{L}_2}}$)
in Table \ref{tab:proposed-results}. 
For comparison, we train a separate autoencoder on the displacement
fields produced by UnDR (Table \ref{tab:proposed-results}). 
These results are in agreement with the key idea that the CAE helps the primary network to produce results closer to a low-dimensional manifold, as
represented by the ability of the bottle-neck AE to accurately reconstruct its output. 

\begin{table}[!h]
\centering
\vspace{-10pt}
\begin{tabular}{|l|l|r|r|r|r|}
\hline
\textbf{Dataset} \hspace{10mm} & \textbf{Method} &  \textbf{AE error} &  \textbf{Dice coeff.} & \textbf{Landmark error} & \textbf{Test runtime}\\ \hline
Linear Box-Bump & CAE (1, $\beta$ = 8) & 6.8\% & 0.98 & 26\%  & 0.0185s \\ \hline
Linear Box-Bump & UnDR & 66.4\% & 0.97 & 124\% & 0.0184s \\ \hline
Linear Box-Bump & UnDR-BN & 65.8\% & 0.96 & 122\% & 0.0190s \\ \hline
\hline
Rotate Box-Bump & CAE (1, $\beta$ = 8) & 12.3\% & 0.98 & 24\%  & 0.0195s \\ \hline
Rotate Box-Bump & UnDR  & 63.5\%  & 0.99 & 101\% & 0.0193s \\ \hline
Rotate Box-Bump & UnDR-BN & 54.9\% & 0.99 & 102\% & 0.0196s \\ \hline
\hline
Corpus Callosum & CAE (2, $\beta$ = 10)  & 33.2\% & 0.89 & 5.7 mm  & 0.0237s \\ \hline
Corpus Callosum & CAE (4, $\beta$ = 10)  & 19.2\% & 0.93 & 5.1 mm  & 0.0237s \\ \hline
Corpus Callosum & CAE (8, $\beta$ = 10)  & 18.5\% & 0.95 & 4.5 mm  & 0.0237s \\ \hline
Corpus Callosum & CAE (16, $\beta$ = 10)  & 16.3\% & 0.96 & 5.2 mm  & 0.0237s \\ \hline
Corpus Callosum & UnDR & 33 - 63\%${^\dagger}$ & 0.93 & 6.5 mm  & 0.0234s \\ \hline
\hline
Left Atrium (3D) & CAE (5, $\beta$ = 0.2) & 29.8\% & 0.76  & 9.9 mm & 0.784s\\ \hline
Left Atrium (3D) & UnDR & 46.3\%  & 0.75 & 10.1 mm & 0.772s \\ \hline
\end{tabular}
\vspace{5pt}
\caption{Results obtained with Cooperative AutoEncoder networks (CAE, bottleneck size, $\beta$ coefficient) compared with Unsupervised Deformable Registration (UnDR) by \protect\cite{balakrishnan2018unsupervised}. Landmark errors for box-bump datasets are reported as the percentage of bump width. The AE error for UnDR refers to a separate autoencoder with bottleneck size same as CAE bottleneck (trained after UnDR). $^{\dagger}$ The AE error is 63.3\% for bottleneck size 1, 54.1\% for 2, 49.4\% for 4, 38.8\% for 8, and 33.5\% for 16. We also report the average test runtime to compute the displacement fields.
}
\label{tab:proposed-results}
\vspace{-15pt}
\end{table}


\begin{figure}[!t]
    \centering
    \vspace{-5pt}
    \includegraphics[width=0.85\textwidth]{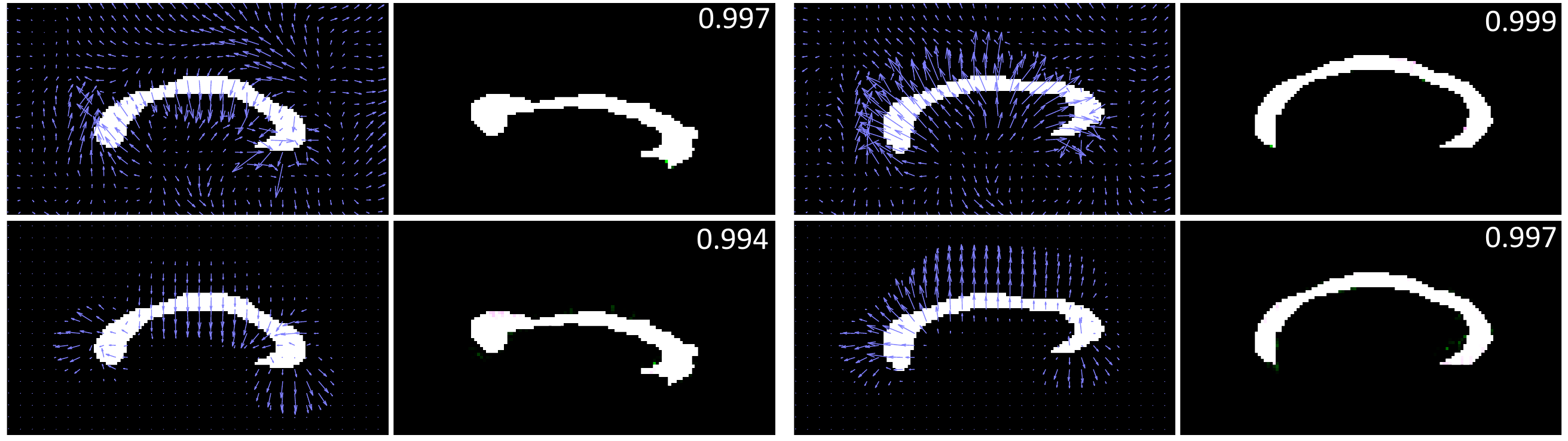}
    \vspace{-5pt}
    \caption{Two corpus callosum source-target pairs, again one image showing the fields and the other a falsecolor between target and the registered output; top-row: UnDR,
      bottom-row: CAE.}
    \label{fig:my_label}
    \vspace{-20pt}
\end{figure}

\noindent\textbf{Corpus Callosum (CC)}

In this example, we use a dataset of 324 mid-saggital 2D slices of Corpus Callosum (CC) from the OASIS Brains dataset \cite{oasis2010}.  Unlike synthetic
experiments discussed above, we do not know, apriori, the intrinsic
dimensionality of the CC shapes.  Therefore, we
train the proposed architecture across a range of CAE bottleneck dimensions
(2, 4, 8 and 16) and compare resulting Dice coefficients, autoencoder reconstructions, and landmark errors, as in Table
\ref{tab:proposed-results}. Networks are again trained using $\mathbb{L}_2$ difference as the primary loss. Landmarks were identified using features
from the literature \cite{sigirli2012shape}, and we had multiple raters identify the posterior
and anterior points of the CC, the inferior tip of the splenium, the
posterior tip of the genu, the posterior angle of the genu, and the
interior notch of the splenium.  Interrater RMS error is 1.4mm, and
the pixel/voxel size is 1mm for these images.  We see that the optimal
bottleneck size for cooperative networks is 8 -- increasing the
bottleneck to 16 improves the Dice coefficient and AE error, but leads
to worse landmark error, which suggests the CAE 
starts to overfit.
The UnDR approach leads to comparable Dice
scores, but worse autoencoder and landmark errors (Table
\ref{tab:proposed-results}). As in the synthetic experiments, to
report the AE error for UnDR, we trained the autoencoder separately
after UnDR training. CAE helps the primary network produce displacement fields that are close to a low-dimensional manifold---a result that is not achieved with the conventional smoothness penalty.

\vspace{2pt}
\noindent\textbf{Left Atrium Appendage (LAA)}

\setlength{\intextsep}{0pt}
\setlength{\columnsep}{5pt}
\begin{wrapfigure}[7]{r}{2.6in} 
  \begin{center}
    \includegraphics[scale = 0.40, trim = 5mm 0mm 0mm 35mm]{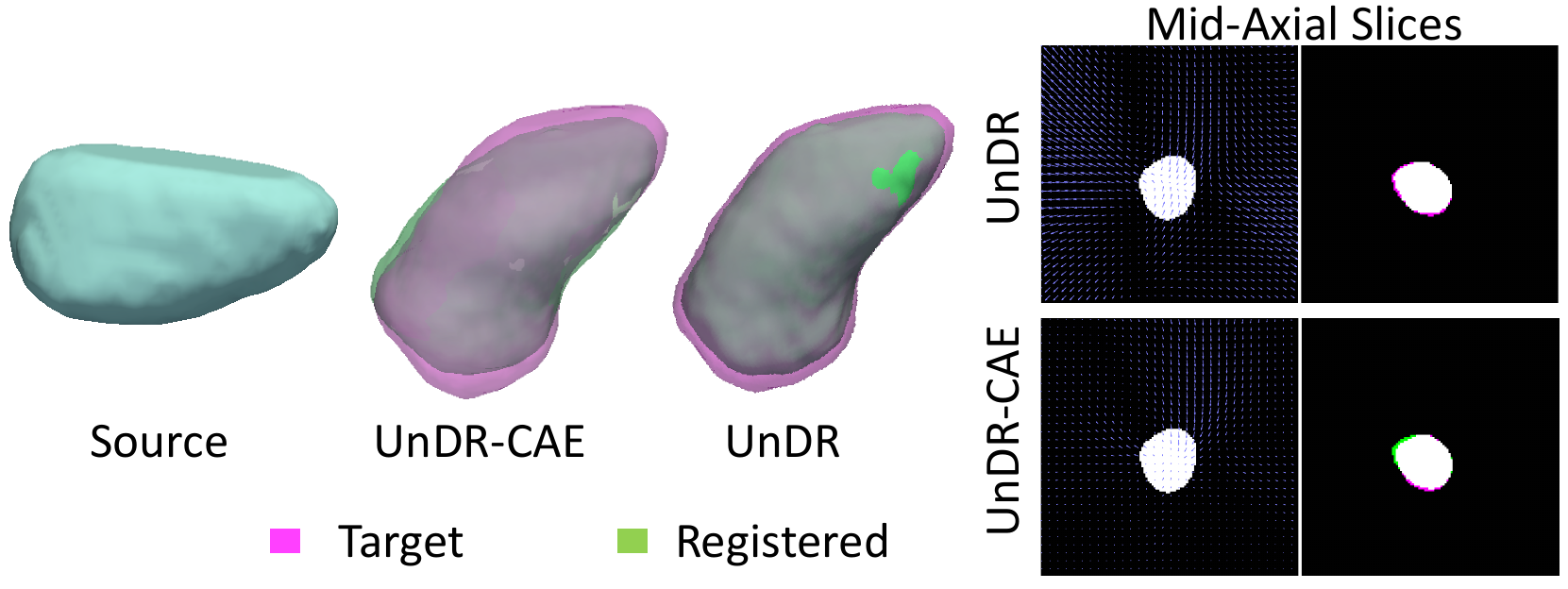}
    \vspace{-19pt}
    \caption{The results of the 3D LAA registration produced by cooperative networks and UnDR.}
    \label{fig:LAA}
  \end{center}
\end{wrapfigure}
\setlength{\intextsep}{12pt plus 2.0pt minus 2.0pt}
\setlength{\columnsep}{10pt}

We apply the cooperative network on a 3D dataset of left atrium appendages
(LAA). These images are represented as signed distance transforms, and
hence we use the normalized cross-correlation loss as in
\cite{balakrishnan2018unsupervised}, instead of a $\mathbb{L}_2$ image loss. The Dice scores, AE
reconstruction accuracy and compute times are reported in Table \ref{tab:proposed-results}. We also show the registration of a pair of LAA images in Figure \ref{fig:LAA}, and landmark (manually obtained clinically validated Ostia landmarks on LAA) reconstruction errors in Table \ref{tab:proposed-results}.

\vspace{-15pt}
\section{Conclusions}
\vspace{-10pt}
This paper proposes a novel architecture proposed for CNN-based unsupervised image registration that uses a cooperative autoencoder (CAE) and enforces the displacement fields to lie in the vicinity of a low-dimensional manifold. CAE reconstruction loss acts as a regularizer term for unsupervised registration. Cooperative networks have comparable registration run times (Table \ref{tab:proposed-results}) with UnDR, but much faster as compared to the conventional state-of-the-art registration methods (as analyzed in \cite{balakrishnan2018unsupervised}). Cooperative networks produce meaningful correspondence representation between shapes as compared to other methods (evident by landmark reconstruction errors in Table \ref{tab:proposed-results}), while maintaining the registration accuracy, making it a viable tool for obtaining fast alignment with anatomically feasible correspondence.

\textbf{Acknowledgements : } This work was supported by NIH [grant numbers R01-AR-076120-01, R01- HL135568-02, and P41-GM103545-19] and  also supported by the National Institute of General Medical Sciences of the National Institutes of Health under grant number P41 GM103545-18.
\appendix

\bibliographystyle{splncs03}
\vspace{-2pt}

\bibliography{main}
\section*{Appendix}
\section{UnDR-BN Without Skip-Connections}
\label{app:undr-bnwosc}
\begin{figure}
    \centering
    \includegraphics[width=\textwidth]{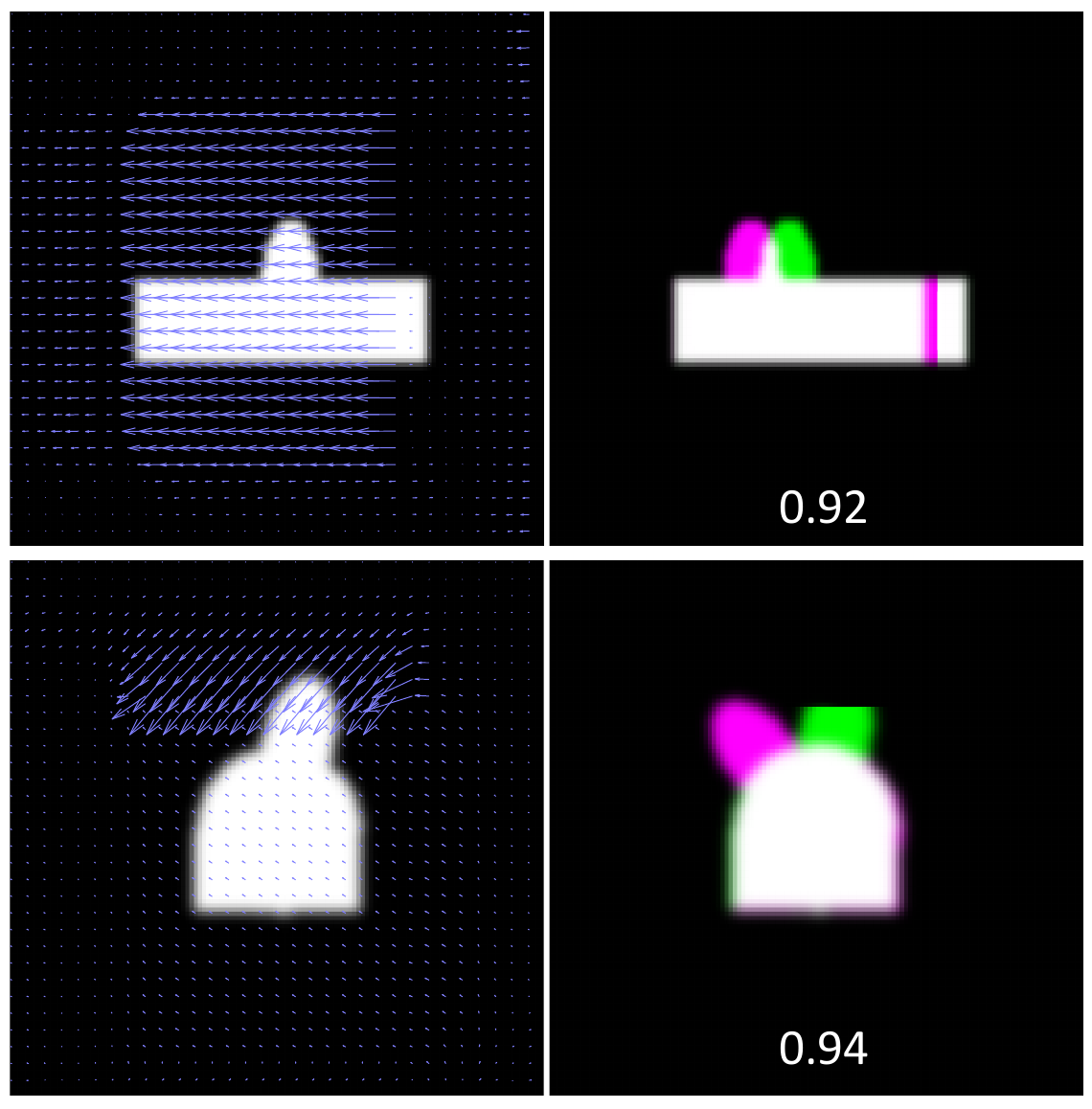}
    \caption{\emph{Top-row:} an example pair of liner box-bump images from testing set passed through UnDR-BN without skip-connections. left image shows the source image with the displacement fields. Right image shows difference image between the registered output and the true target image, the white pixels are correctly matched and the green and magenta shows the mismatched pixels of registered and target images respectively. \emph{Bottom-row: } an example from the rotating box-bump dataset.}
    \label{fig:skipconn}
\end{figure}

In the paper we consider comparison of CAE based UnDR with the variant of a constrained bottleneck, however, we saw similar results which we attributed to the still in place \emph{skip-connections} of the U-Net architecture. Here, we redo the experiments of a single dimensional bottleneck in primary U-Net architecture for linear and rotating box-bump datasets, but we remove the skip-connections. This architecture is essentially a hard constraint on the dimensionality of displacement field as compared to the soft constraint as imposed by the CAE. In Figure \ref{fig:skipconn} we show the results for these experiments. We can see that the registration is subpar, which aligns with our assumption that a CAE based soft penalty, to lie close to a low-dimensional manifold produces more accurate registrations and corresponding flow-fields as compared to hard penalty for the population of displacement fields to lie exactly on the low-dimensional manifold. The average dice coefficient for linear box-bump with this method is 0.926 and for rotating box-bump is 0.974.

\section{Latent Space Variations}
\label{app:latentspace}
\begin{figure}
    \centering
    \includegraphics[width=\textwidth]{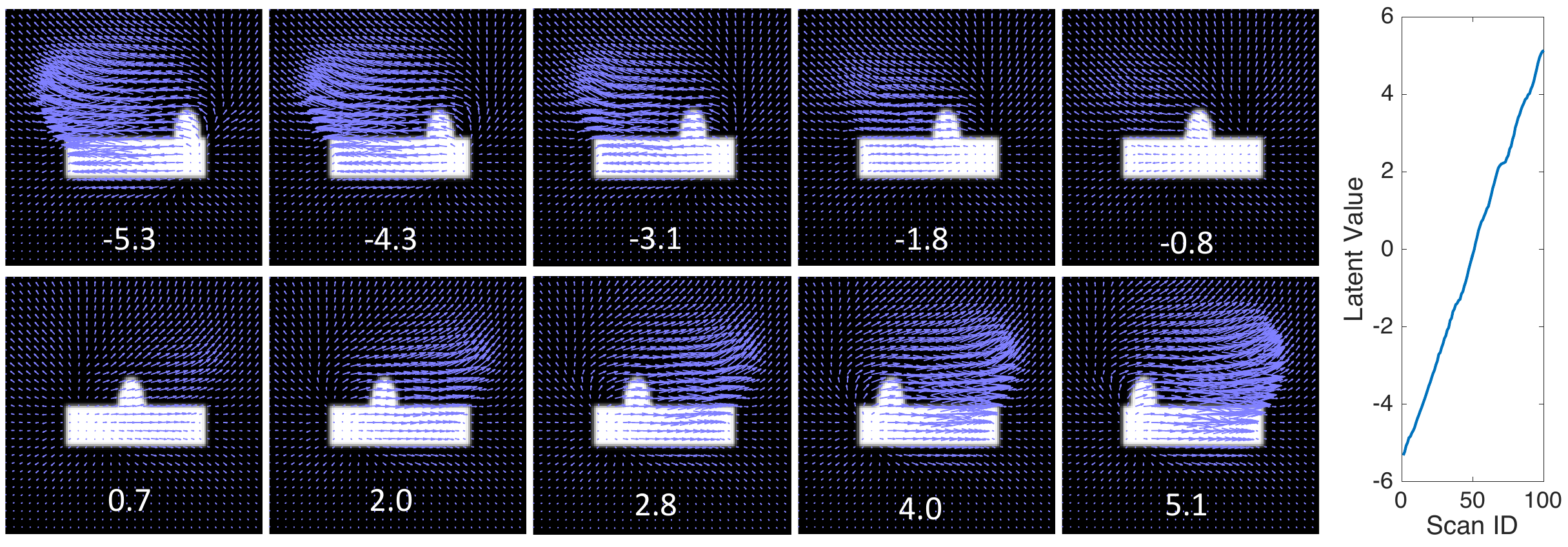}
    \caption{\emph{Linear Box-Bump Data: } The images going from top to bottom , left to right represent the 10 uniform samples of the source images and the corresponding displacement fields registering them to central bump image, the numbers in white denote the value of the latent space of the CAE. The plot on the right denotes the latent space value of the CAE for all 100 source images when registered to the central bump image.}
    \label{fig:linearLatentVar}
\end{figure}

\begin{figure}
    \centering
    \includegraphics[width=\textwidth]{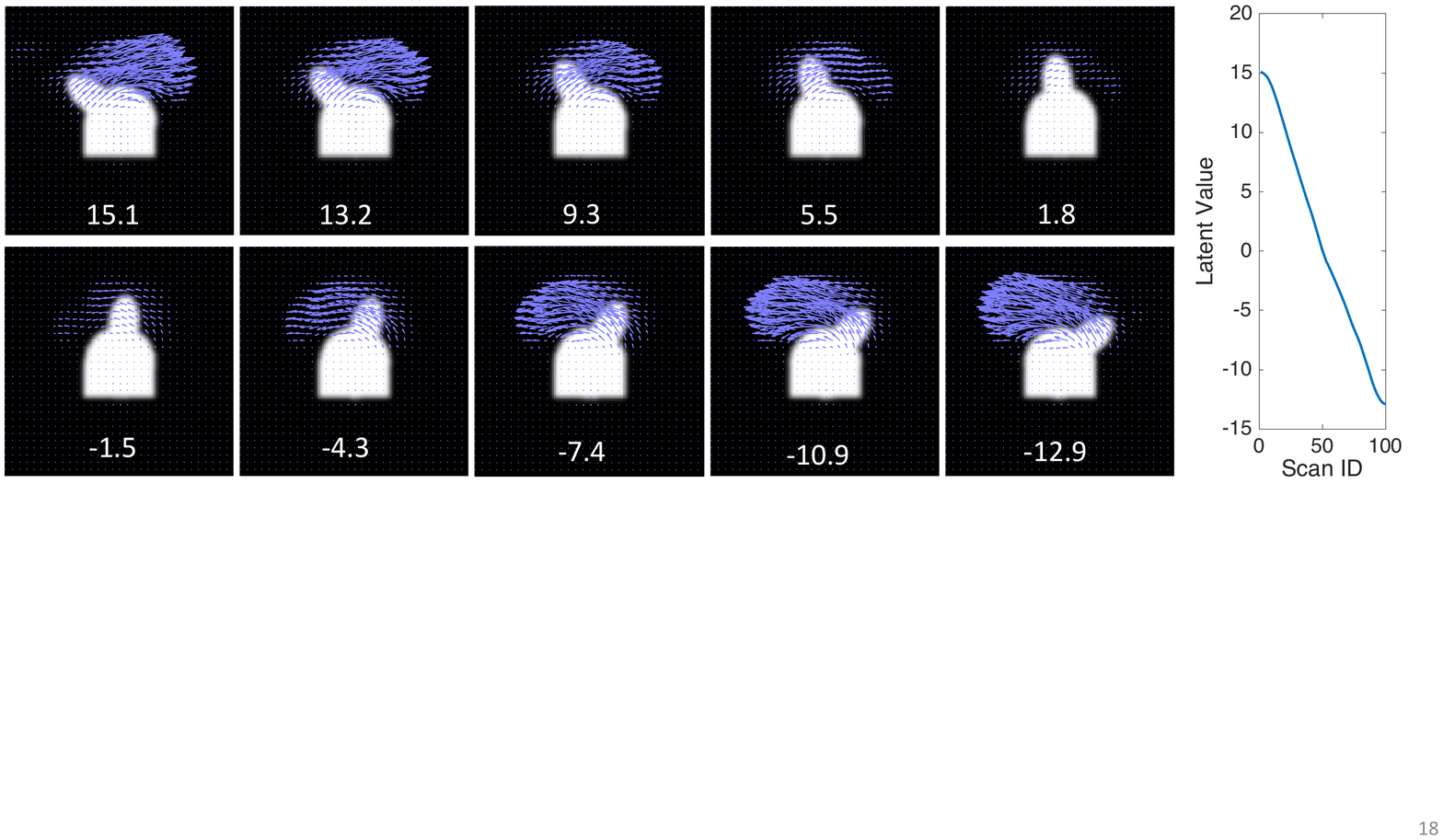}
    \caption{\emph{Rotating Box-Bump Data: } The images going from top to bottom , left to right represent the 10 uniform samples of the source images and the corresponding displacement fields registering them to central bump image, the numbers in white denote the value of the latent space of the CAE. The plot on the right denotes the latent space value of the CAE for all 100 source images when registered to the central bump image.}
    \label{fig:rotateLatentVar}
\end{figure}

We examine the nature of the latent space of the CAE for the linear and rotating box-bump datasets. For each dataset, we fix the target image to be the mean image (each in both case is the image with the bump in center) and change the source images. For linear box-bump we have 100 source images and in Figure \ref{fig:linearLatentVar}(images) we show 10 equally spaced source images with their corresponding displacement field bringing each image to the center bump target image. Figure \ref{fig:linearLatentVar}(plot) shows the value of the CAE bottleneck versus the source images. The variation in the latent space is monotonic and makes intuitive sense corresponding to displacement field it generates. This behavior is same for the rotating box bumps whose variations are seen in Figure \ref{fig:rotateLatentVar}.

\end{document}